\pdfoutput=1

\documentclass[11pt]{article}

\usepackage{ACL2023}
\usepackage{times}
\usepackage{latexsym}
\usepackage{graphicx}
\usepackage{multirow}

\usepackage[T1]{fontenc}

\usepackage[utf8]{inputenc}

\usepackage{microtype}

\usepackage{inconsolata}

\title{ArgU: A Controllable Factual Argument Generator}

\author{Sougata Saha and Rohini Srihari\\
State University of New York at Buffalo\\
Department of Computer Science and Engineering\\
\texttt{\{sougatas, rohini\}@buffalo.edu}}

\begin{document}
\maketitle
\begin{abstract}

Effective argumentation is essential towards a purposeful conversation with a satisfactory outcome. For example, persuading someone to reconsider smoking might involve empathetic, well founded arguments based on facts and expert opinions about its ill-effects and the consequences on one's family. However, the automatic generation of high-quality factual arguments can be challenging. Addressing existing controllability issues can make the recent advances in computational models for argument generation a potential solution. In this paper, we introduce ArgU: a neural argument generator capable of producing factual arguments from input facts and real-world concepts that can be explicitly controlled for stance and argument structure using Walton's argument scheme-based control codes. Unfortunately, computational argument generation is a relatively new field and lacks datasets conducive to training. Hence, we have compiled and released an annotated corpora of 69,428 arguments spanning six topics and six argument schemes, making it the largest publicly available corpus for identifying argument schemes; the paper details our annotation and dataset creation framework. We further experiment with an argument generation strategy that establishes an inference strategy by generating an ``argument template'' before actual argument generation. Our results demonstrate that it is possible to automatically generate diverse arguments exhibiting different inference patterns for the same set of facts by using control codes based on argument schemes and stance.

\end{abstract}

\section{Introduction}

Although arguing is an innate human quality, formulating convincing arguments is an art. A successful  narrative aiming to persuade someone should be rhetorically appealing, trustworthy, factually correct, and logically consistent, which makes formulating good arguments challenging. Incorporating neural language models, the relatively new field of computational argument generation has shown promise in assisting with argument synthesis. Argument generators like Project Debater \cite{project-debater} have successfully formulated convincing arguments across different domains including legal, politics, education, etc., and can potentially find new argumentative connections. However, lacking explicit control mechanisms, neural argument generators often render illogical and inappropriate arguments, reducing their trustworthiness and applicability for practical use. Furthermore, training such models requires a considerable amount of quality data, which is hard to collect and annotate. Hence, we propose ArgU, a controllable neural argument generator trained on a curated and quality-controlled corpus of annotated argument texts from abortion, minimum wage, nuclear energy, gun control, the death penalty and school uniform. 

\begin{figure}[h]
  \centering
  \includegraphics[width=\columnwidth]{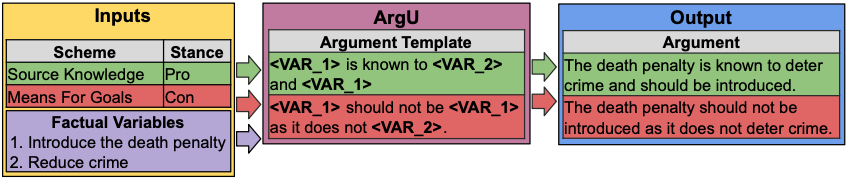}
  \caption{Generating stance and argument scheme controlled factual arguments using ArgU.}
  \label{fig:argu-intro}
\end{figure}

ArgU strives to enable effective, scalable and appealing argument generation. As depicted in Figure \ref{fig:argu-intro}, it takes as input worldly knowledge and concepts as variables and coherently combines them to generate an argument that exhibits the desired pro/con stance and inference structure. Using control codes to regulate argument stance and reasoning, ArgU generates a variety of argument texts for the same set of facts, thus providing diverse response options. Internally ArgU implements a 2-step generation process, where it first generates an ``argument template'', which depicts the structure of the final argument based on the control codes, and finally yields the argument text by modifying the template to include the augmented input fact variables. We ground our work on prominent theoretical foundations, where the inference structure-based control codes derive from six Walton's argument schemes: ``Means for Goal'', ``Goal from Means'', ``From Consequence'', ``Source Knowledge'', ``Source Authority'', and ``Rule or Principle''. 

Since human annotation is expensive and time-consuming, we devise a multi-phased annotation framework for systematically leveraging human and automatic annotation mechanisms to yield a curated dataset of 69,428 examples for controllable argument synthesis. We release our curated corpus to facilitate further research;  an example constitutes an argument text, a set of real-world concepts and knowledge from which the argument derives, and the stance and argument scheme of the text. We further detail and  analyze our annotation framework and share variants of topic-independent computational models for automatically annotating factual spans from argument text and identifying the asserted argument schemes. We summarize our contributions below:
\begin{itemize}
    \item We propose an argument generator that methodically generates factual arguments following a specified stance and argument scheme (Sec. \ref{argu-gen}).
    
    \item We share a quality-controlled annotated dataset conducive to training such generators. To our knowledge, this is the largest available corpora that identify argument schemes from argument text (Sec. \ref{final-dataset}).
    
    \item We share our annotation framework and release domain-independent computational models that automatically identify factual spans and argument schemes from argument text from any topic (Sec. \ref{annotation-framework}).
\end{itemize}



\section{Related Work}
Argument schemes are typical inference patterns found in arguments. Walton provided an in-depth study of argument schemes \cite{walton2008argumentation} and defined 60 such schemes prevalent in daily argument text. Based on Walton's argumentation schemes, \citet{kondo-etal-2021-bayesian} proposed representing the reasoning structure of arguments using Bayesian networks and defined abstract network fragments termed idioms, which we use here.

Advances in neural methods for language modelling have enabled the field of computational argument generation. \citet{hua-wang-2018-neural} introduced a factual argument generator that generates opposite stance arguments by yielding a set of talking point key phrases, followed by a separate decoder to produce the final argument text. \citet{hua-etal-2019-argument-generation} proposed Candela, a framework for counter-argument generation similar to \citet{hua-wang-2018-neural}, which also controls for the style. \citet{aspect_controlled} introduced Arg-CTRL: a language model for generating sentence-level arguments using topic, stance, and aspect-based control codes \cite{keskarCTRL2019}. \citet{Khatib2021EmployingAK} constructed argumentation-related knowledge graphs and experimented with using them to control argument generation. \citet{Alshomary2021CounterArgumentGB} explored a novel pipelined approach to generating counter-arguments that first identifies a weak premise and then attacks it with a neurally generated counter-argument. Hypothesizing that the impact of an argument is strongly affected by prior beliefs and morals, \citet{alshomary-etal-2022-moral} studied the feasibility of the automatic generation of morally framed argument text and proposed an argument generator that follows the moral foundation theory. \citet{Syed2021GeneratingIC} introduced the task of generating informative conclusions from arguments. They compiled argument text and conclusion pairs and experimented with extractive and abstractive models for conclusion generation using control codes. \citet{Chakrabarty2021ENTRUSTAR} experimented with argument text re-framing for positive effects. They created a suitable corpus and trained a controllable generator with a post-decoding entailment component for re-framing polarizing and fearful arguments such that it can reduce the fear quotient. Our work best aligns with Arg-CTRL and Candela, where we use control codes to regulate argument generation and implement a multi-step decoding pipeline to generate the final argument. However, unlike Arg-CTRL, we control for the argument scheme, and unlike Candela, our multi-step decoding utilizes an argument template as an intermediate step.


Most argumentation datasets identify argumentative components (claims, premises, etc.), making them better suited for argument-mining tasks \cite{stab-gurevych-2014-annotating, Peldszus2015AnAC, ghosh-etal-2016-coarse, hidey-etal-2017-analyzing, chakrabarty-etal-2019-ampersand}. Further, existing argument scheme annotated corpora are either very restricted in domain and size \cite{reed-etal-2008-language, feng-hirst-2011-classifying, green2015identifying, musi2016towards, visser2022annotating, jo2021classifying} or only provide guidelines and tools for annotations \cite{visser2018revisiting, lawrence2019online}. Hence, we use the BASN dataset \cite{kondo-etal-2021-bayesian}, which contains sizeable examples spanning six topics and identify argument schemes.

\section{Argument Generation Corpus}
\label{annotation-framework}
Training a factual argument generator controlled for the stance and argument scheme requires examples that identify such features from the text: such a corpus is lacking. Hence, we introduce a two-phased annotation framework that yields a corpus of 69,428 examples which (i) identify argument schemes and factual spans from argument text and (ii) grounds the spans to a knowledge base (KB). In the first phase, we employ human annotators to identify factual spans from a subset of an existing dataset of 2,990 arguments which already identifies argument schemes. We further train computational models to annotate the remaining corpus for factual spans and perform extensive quality checks. In the second phase, we train models from the resultant Phase 1 dataset to automatically annotate a larger parallel corpus for both argument scheme and factual spans, yielding an annotated corpus\footnote{All dataset and code to be released post acceptance.} of 69,428 arguments for training argument generators.  

\subsection{Phase 1 (P1): Initial Corpus Creation} 
\citet{kondo-etal-2021-bayesian} introduced the BASN dataset comprising 2,990 pairs of arguments and abstract network fragments derived from six Walton's argumentation schemes: ``Means for Goal'', ``Goal from Means'', ``From Consequence'', ``Source Knowledge'', ``Source Authority'', ``Rule or Principle'', and ``Others''. They utilized a knowledge base (KB) of 205 facts (termed as variables) spanning the topics of abortion, minimum wage, nuclear energy, gun control, the death penalty and school uniform to define the idioms. Figure \ref{fig:phase_1} (Appendix \ref{sec:appendix}) illustrates an example from the BASN dataset where variables from the KB formulate a pro-stance argument following the ``Means for goals'' argument scheme. We perform two annotation tasks in P1: (i) \textbf{Span Detection}: Annotate arguments by identifying (highlighting) non-overlapping factual spans from argument text. (ii) \textbf{Span Grounding}: Ground the identified factual spans to the available KB variables, or ``Others'' if the span is unrelated to any available variables.


We annotate 1,153 randomly sampled examples spanning all six topics and train a model for automatically annotating the remaining examples. We further perform human evaluations to determine the correctness of the automatic annotations. 


\subsubsection{Human Expert Annotation}
\label{human-annotators}
Using Doccano \cite{doccano}, we annotated 1,153 examples from the BASN corpus for both the tasks of span detection and grounding, where each sample comprised an argument and a minimum of 2 to a maximum of 5 fact variables from the KB. Figure \ref{fig:doccano_screenshot} (Appendix \ref{sec:appendix}) contains a screenshot from our Doccano annotation task. We employed two expert annotators with a background in computational linguistics and computer science for the annotation task. To be efficient with resources, each annotator independently annotated non-overlapping examples. Further, to ensure consistency across annotations, we computed inter-annotator agreement over 66 samples, which resulted in a Cohen's Kappa score of 0.79, indicating substantially high agreement.

\subsubsection{Automatic Annotation: ArgSpan}

\begin{figure}[h]
  \centering
  \includegraphics[width=\columnwidth]{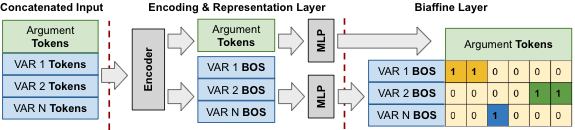}
  \caption{ArgSpan Architecture.}
  \label{fig:ArgSpan-architecture}
\end{figure}

We train ArgSpan: a Roberta-based tagger \cite{liu2019roberta}, on the annotated examples for automatically annotating the rest of the BASN dataset for both tasks. Figure \ref{fig:ArgSpan-architecture} illustrates ArgSpan's architecture. ArgSpan inputs concatenated argument and fact variables and encodes them using a Roberta-based encoder. It reduces the hidden representation for each fact variable by passing the beginning of the string token (BOS) through a fully connected neural network layer. Finally, it uses a biaffine layer to capture the interaction between the argument text and each variable. The model is trained end-to-end by minimizing the cross entropy loss between the predicted logit for each argument token and the actual BIO scheme encoded target label. Appendix \ref{arg-span-details} contains further training details.

\subsubsection{Evaluation}


We automatically annotate the remaining BASN samples using ArgSpan. To gauge the quality of the automatic annotations, we ask one of the human evaluators to annotate 300 random examples from the remaining samples using Doccano and compare them with the model predictions. Detailed in Figure \ref{fig:ArgSpan-eval} (Appendix \ref{sec:appendix}), we evaluate Span Detection by computing the F1 score between the overlapping predicted and human-identified tokens and achieve an average score of 91.1\% across all 300 examples. We measure accuracy for evaluating Span Grounding and attain a score of 89.2\%. With the additional 300 examples (total of 1,453), we re-train ArgSpan and perform inference on the remaining BASN samples, yielding a fully annotated corpus of 2,990 examples with KB-grounded factual spans and argument schemes from argument text. Also, we observe very few examples of the ``Goal From Means'' scheme in the resultant dataset and combine it with the more prevalent ``Means for Goal'' scheme, resulting in six argument schemes.


\subsection{Phase 2 (P2): Corpus Expansion}

\citet{kondo-etal-2021-bayesian} used crowd-sourcing to create the BASN dataset, where crowd workers formulated argument text from a knowledge base comprising a limited number of premise-conclusion pairs (fact variables). Although such an approach resulted in a considerable number of arguments, using approximately 34 fact variables per topic, it lacks variety. Training an argument generator on such a corpus would limit its generalizability and use. Hence, we expand the P1 dataset with a parallel corpus (\textbf{PC}) of 66,180 examples from the Aspect-Controlled Reddit and CommonCrawl corpus by \citet{aspect_controlled}, and 733 combined examples from the Sentential Argument Mining, Arguments to Key Points and the debate portal-based Webis datasets \cite{sentential_am, arg_kp1, arg_kp2, webis}. Since the PC examples do not identify factual spans and argument schemes, we use the fully annotated P1 dataset to train \textbf{ArgSpanScheme}: a Roberta-based model that identifies factual spans and argumentation schemes from argument text. We automatically annotate the PC using ArgSpanScheme and combine them with the P1 dataset, to yield the P2 dataset. 


\subsubsection{ArgSpanScheme Architecture}
\begin{figure}[h]
  \centering
  \includegraphics[width=\columnwidth]{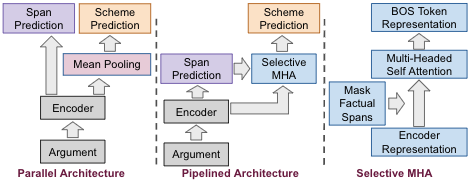}
  \caption{ArgSpanScheme Architectures.}
  \label{fig:ArgSpanScheme-architectures}
\end{figure}

Illustrated in Figure \ref{fig:ArgSpanScheme-architectures}, we experiment with two variants of ArgSpanScheme to jointly extract factual spans and predict argument schemes from argument text. Both architectures use a Roberta-based encoder to encode an input argument text and differ in the final prediction layers, as detailed below.\\
\noindent
\textbf{Parallel Architecture}
Here we use two independent classification heads: (i) A span detection head which uses a linear layer to extract factual spans by classifying each encoded argument token as belonging to one of the three BIO tags. (ii) A scheme detection head which uses a linear layer to predict argument schemes by performing a multi-label (six labels including ``Others'') classification on the mean pooled encoded argument tokens.\\
\noindent
\textbf{Pipelined Architecture}
Argument schemes represent structures of inference and are invariant to the constituent facts. For example, although both arguments A: ``Increase in the minimum wage is not favourable as it can increase unemployment'', and B: ``Increase in gun laws are favourable as it reduces gun violence'', are from different topics, they follow a similar structure ``X is/are (not) favourable as it Y'', exhibiting ``From Consequences'' argument scheme. As depicted in Figure \ref{fig:ArgSpanScheme-architectures}, we model this by performing selective multi-headed attention. We mask the factual spans predicted by the span detection head and apply two layers of multi-headed self-attention on the remaining tokens. Finally, we pass the BOS token representation through a linear layer to predict the argument schemes. Appendix \ref{arg-span-scheme-details} contains further training details.

\subsubsection{Modelling Results and Evaluation}


\begin{table*}[t]
\centering
\resizebox{\textwidth}{!}{%
\begin{tabular}{c|ccc|ccccccc}
\hline
\multicolumn{1}{c|}{\textbf{}} &
  \multicolumn{3}{c|}{\textbf{Span}} &
  \multicolumn{7}{c}{\textbf{Scheme}} \\ \hline
\multicolumn{1}{c|}{\textbf{Split}} &
  \multicolumn{1}{c}{\textbf{Partial}} &
  \multicolumn{1}{c|}{\textbf{Full}} &
  \multicolumn{1}{c|}{\textbf{Overall}} &
  \multicolumn{1}{c}{\textbf{\begin{tabular}[c]{@{}c@{}}From \\ Consequence\end{tabular}}} &
  \multicolumn{1}{c}{\textbf{\begin{tabular}[c]{@{}c@{}}From Source \\ Authority\end{tabular}}} &
  \multicolumn{1}{c}{\textbf{\begin{tabular}[c]{@{}c@{}}From Source \\ Knowledge\end{tabular}}} &
  \multicolumn{1}{c}{\textbf{\begin{tabular}[c]{@{}c@{}}Goal From Means/\\ Means from Goals\end{tabular}}} &
  \multicolumn{1}{c}{\textbf{\begin{tabular}[c]{@{}c@{}}Rule or \\ Principle\end{tabular}}} &
  \multicolumn{1}{c|}{\textbf{Other}} &
  \multicolumn{1}{c}{\textbf{Overall}} \\ \hline
CV &
  \textbf{0.86}/0.85 &
  \multicolumn{1}{l|}{\textbf{0.92}/0.91} &
  0.89/0.89 &
  \textbf{0.94}/0.93 &
  \textbf{0.92}/0.91 &
  0.88/\textbf{0.90} &
  \textbf{0.96}/0.95 &
  \textbf{0.97}/0.96 &
  \multicolumn{1}{l|}{\textbf{0.88}/0.86} &
  \textbf{0.95}/0.94 \\
5:1 &
  0.70/0.70 &
  \multicolumn{1}{l|}{0.77/\textbf{0.78}} &
  0.81/0.81 &
  0.65/0.65 &
  0.68/\textbf{0.85} &
  0.48/0.48 &
  0.48/\textbf{0.56} &
  0.64/\textbf{0.66} &
  \multicolumn{1}{l|}{0.46/0.46} &
  0.68/\textbf{0.69} \\
4:2 &
  0.76/\textbf{0.77} &
  \multicolumn{1}{l|}{0.84/0.84} &
  0.85/0.85 &
  0.60/\textbf{0.71} &
  0.67/\textbf{0.70} &
  0.49/0.49 &
  0.45/\textbf{0.47} &
  0.49/\textbf{0.55} &
  \multicolumn{1}{l|}{0.49/0.49} &
  0.75/\textbf{0.82} \\
2:4 &
  0.74/0.74 &
  \multicolumn{1}{l|}{\textbf{0.82}/0.80} &
  \textbf{0.82}/0.80 &
  0.63/\textbf{0.73} &
  \textbf{0.69}/0.67 &
  \textbf{0.50}/0.49 &
  \textbf{0.47}/0.46 &
  0.73/\textbf{0.77} &
  \multicolumn{1}{l|}{0.46/0.46} &
  0.70/\textbf{0.77} \\ \hline
\end{tabular}%
}
\caption{ArgSpanScheme span and scheme prediction results for Parallel / Pipelined versions. The best performing model for each data split and task is highlighted in bold.}
\label{tab:m2-scheme-results}
\end{table*}

For both tasks of span and scheme detection, we compare the F1 score of the parallel and pipelined architectures across different data splits. We perform a 5-fold Cross Validation (CV) by randomly splitting the resultant dataset from P1 into 93\% training and 7\% validation split. We further assess the generalizability of ArgSpanScheme by training and validating on examples from non-overlapping topics. As illustrated in Figure \ref{fig:ArgSpanScheme-data} (Appendix \ref{sec:appendix}), we set up five data splits (ids 1 to 5) comprising three combination ratios of training-validation topics (5:1, 4:2, and 2:4), which increases the difficulty by reducing the number of training topics. 

\noindent
\textbf{Evaluating Span Prediction:} For span detection we compute the F1 score at three levels of overlap: (i) \textbf{Partial Overlap}: A span level metric where a predicted span is true positive if at least 50\% of its tokens overlap with the actual span. (ii) \textbf{Full Overlap}: A span level metric where a predicted span is true positive if all of its tokens overlap with the actual span. (iii) \textbf{Overall}: A token level metric which compares the predicted and actual token BIO labels. Table \ref{tab:m2-scheme-results} shares the CV and combination ratio aggregated results for span detection. We observe similar performance for both ArgSpanScheme versions across all three levels of overlap.


\noindent
\textbf{Evaluating Scheme Prediction:} We compare scheme-wise and overall F1 scores and share the results in Table \ref{tab:m2-scheme-results}. We observe that the parallel architecture slightly outperforms the pipelined version in CV, whereas the pipelined version almost always performs better for the non-overlapping splits. The results indicate that for scheme detection, incorporating a generalizable architecture by emphasizing the argument structure rather than the factual spans does lead to better results on unseen topics.

\noindent
\subsubsection{Automatic Annotation \& Human Eval.} 
Based on the analysis of automatic evaluation results, we train a final pipelined version of ArgSpanScheme on the P1 dataset and perform inference on the PC to automatically annotate it for factual spans and argument schemes. We randomly sample 200 annotations and perform a human evaluation using one evaluator to ascertain the annotation quality. 

\noindent
\textbf{Evaluating Span Prediction:} We present the human evaluator with an argument text along with the model predicted spans and ask them to rate each example using two custom metrics: (i) \textbf{Span Precision}: On a continuous scale of 1 (low) to 5 (high), how sensible are the identified spans? Spans which are unnecessarily long or abruptly short are penalized. This metric evaluates whether the identified spans adequately convey meaningful information. (ii) \textbf{Span Recall}: On a continuous scale of 1 (low) to 5 (high), how well does the model perform in identifying all factual spans? Examples which fail to identify spans conveying real-world concepts and factual knowledge are penalized. We observe an average score of 4.1 (median 4.7) for Span Precision and 3.9 (median 4.4) for Span Recall, indicating the reliability of the automatic annotations.

\noindent
\textbf{Evaluating Scheme Prediction:} Since identifying argument schemes is a much more difficult task, we first measure the evaluator's competency by presenting 30 random arguments from the BASN dataset and asking them to label each argument text with the most likely argument scheme. We compared the evaluator-assigned labels with the golden labels and found them to be matching in 53.3\% of cases, with most matches belonging to the ``from consequences'', ``rule or principle'', and ``means for goal'' schemes. Although the labels majorly confirm, the fair amount of disagreement testifies to the task difficulty. Further, Table \ref{tab:annotator-scheme-conflict} (Appendix \ref{sec:appendix}) lists a few examples where we believe the evaluator labels are more accurate than the actual ones. Post-assessment, we asked the evaluator to evaluate the predicted argument schemes of the previously sampled 200 examples with a binary flag, where 1 signifies agreement and 0 signifies disagreement, and observe a fair agreement rate of 73\%.

\subsubsection{Dataset Post-processing}
\label{final-dataset}
The PC initially contains 1,272,548 examples, which we automatically annotate for span and argument scheme using ArgSpanScheme. We persist samples where an argument scheme's predicted probability is at least 20\% of the scheme's average probability and discard examples with the scheme predicted as ``Others''.

To make the PC consistent with the P1 data, we implement the following steps to normalize and ground the ArgSpanScheme-identified factual spans to the existing KB comprising fact variables from BASN or expand the KB with new knowledge wherever applicable. (i) \textbf{Direct Mapping}: Using sentence transformer embedding-based cosine similarity \cite{reimers-2019-sentence-bert} and a threshold of 0.85, we associate factual spans from the annotated PC with its most similar fact variable from the KB. (ii) \textbf{Indirect Mapping}: We use the sentence transformer-based community detection clustering algorithm to cluster similar factual spans from the annotated PC. For directly unmapped spans, we associate the KB fact variable of the nearest neighbour in its cluster. Figure \ref{fig:fact_normalization} (Appendix \ref{sec:appendix}) further illustrates each step in detail.

We apply a series of filtering steps to ensure the quality of the final corpus. We only keep examples containing a maximum of 30\% unnormalized factual spans and add those facts to the KB. Next, we discard instances containing more than 150 words in the argument text and persist examples containing 1-4 fact variables, with each variable present 2-4 times. Finally, to ensure argumentativeness, we parse the argument text using the Dialo-AP argument parser \cite{saha-etal-2022-dialo} and keep examples containing at least one claim. We combine the filtered PC with the P1 dataset to yield 69,428 examples, which we use for argument generation. 

\section{Controllable Argument Generation}
\label{argu-gen}
Arguments based on similar facts but structured differently might lead to dissimilar consequences by exerting different perlocutionary effects. For example, consider argument A: ``Reproductive rights advocates say enabling access to abortion is important towards reproductive rights'', which exhibits the ``From Source Authority'' argument scheme, and B: ``Access to abortion is important towards reproductive rights'', which expresses ``From Consequence''. Although both arguments share the same view regarding the role of abortion in reproductive rights, backed by reproductive rights advocates who are experts, argument A might lead to a favourable outcome in a situation that demands authority. To assist the formulation of arguments exhibiting heterogeneous viewpoints and reasoning, we experiment with BART-based \cite{lewis-etal-2020-bart} neural argument generators capable of generating factual argument text with distinct stances and argument schemes using control codes.


\begin{figure*}[t]
  \centering
  \includegraphics[width=\linewidth]{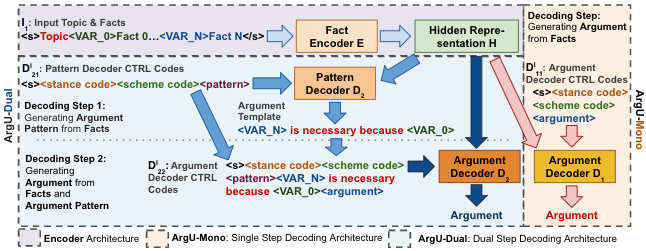}
  \caption{ArgU-Mono and Dual End-to-end Architectures.}
  \label{fig:model_architecture}
\end{figure*}

\subsection{Model Architecture}
Figure \ref{fig:model_architecture} illustrates our encoder-decoder based model architecture, which we discuss below. 
\subsubsection{Encoder}
The model inputs a concatenated representation $\mathrm{I_1}$ of the argument topic and the required KB fact variables. We prefix each variable with a token <VAR\_{X}> where X $\in$ [0, 3] is an incremental id enforcing a random ordering over the variables. The representation $\mathrm{I_1}$ is passed through a BART encoder E to yield a hidden representation H.

\subsubsection{Decoder} A BART based decoder inputs H along with a set of control codes to generate the final argument A. We experiment with two types of decoding:\\
\noindent
\textbf{Single Step Decoding: ArgU-Mono:} As depicted in Figure \ref{fig:model_architecture}, following the standard decoding strategy of an encoder-decoder architecture, the decoder $\mathrm{D_{1}}$ inputs H along with three control codes ($\mathrm{D^{I}_{11}}$) comprising the desired stance, argument scheme, and the argument text BOS token `<argument>', and learns the distribution $\mathrm{P(A|I_1, D^{I}_{11})}$.
\\
\noindent
\textbf{Dual Step Decoding: ArgU-Dual}
An argument generally exhibits structured reasoning by coherently combining variables using appropriate connectives and clauses. For example, the variables A: ``introduce death penalty'' and B: ``reduce crime'' can be combined as ``A \textbf{has shown evidence in} B'', resulting in a pro-death penalty argument ``Introducing the death penalty \textbf{has shown evidence in} reducing crime''. Following the same template of ``A \textbf{has shown evidence in} B'', the variables A: ``enforce gun laws'' and B: ``reduce gun violence'' can be combined to form an argument ``Enforcing gun laws \textbf{has shown evidence in} reduction of gun violence''. The ArgU-Dual architecture implements ``argument templates'' to model this property, where distinct argument texts exhibit similar structure and reasoning over variables. 

To condition the argument generation on its template, we train decoder $\mathrm{D_{2}}$ to create an argument template T before generating the actual argument A. As depicted in Figure \ref{fig:model_architecture}, $\mathrm{D_{2}}$ inputs H and a set of three control codes ($\mathrm{D^{I}_{21}}$) comprising the desired stance, argument scheme, and the template BOS token `<pattern>', to learn the probability distribution $\mathrm{P(T|I_1, D^{I}_{21})}$. Next, we suffix T with the argument BOS token `<argument>`, and pass through $\mathrm{D_{2}}$ to generate the final argument text and learn the distribution $\mathrm{P(A|T, D^{I}_{22})}$.

\subsection{Training, Experiments and Results}
We use the resultant P2 dataset for our experiments and create random train-test set of 67,728 and 1,700 examples. To analyze the effect of each type of control code, we also perform ablation analysis and train two model variants: \textbf{ArgU-Stance} and \textbf{ArgU-Scheme}. Both implementations follow the same encoding and decoding steps as ArgU-Mono, with the only difference being the absence of scheme or stance-based control codes in respective architectures. Training details in Appendix \ref{argu-details}.


\subsubsection{Automatic Evaluation Results}


Apart from comparing standard metrics like corpus BLEU \cite{papineni-etal-2002-bleu} and Rouge-L \cite{lin-2004-rouge}, we define the following metrics to evaluate each model. (i) \textbf{Fact Faithfulness (Fact)}: This evaluates fact faithfulness by measuring the similarity between the input variables and the generated argument. We use the sentence transformer's semantic textual similarity to compute the average cosine similarity between the embeddings of the input variables and the model-generated argument, where a higher score correlates with better utilization of the fact variables. (ii) \textbf{Entailment (Entail) \& Contradiction (Contra)}: This evaluates the relatedness between the original and generated argument. We use AllenNLP's \cite{gardner-etal-2018-allennlp} Roberta-based textual entailment model pre-trained on the SNLI dataset \cite{bowman-etal-2015-large} to determine whether a generated argument entails (higher better) or contradicts (lower better) the original argument with at least 0.8 probability. 

\begin{table}[h]
\resizebox{0.95\columnwidth}{!}{%
\begin{tabular}{r|ccccc}
\hline
\multicolumn{1}{c|}{\textbf{Model}} & \textbf{BLEU} & \textbf{RougeL} & \textbf{Fact} & \textbf{Entail} & \textbf{Contra} \\ \hline
Mono   & 0.150          & 0.379          & 0.641          & 0.399          & 0.140          \\
Dual   & \textbf{0.158} & \textbf{0.381} & 0.641          & \textbf{0.406} & 0.144          \\
Stance & 0.151          & 0.375          & 0.641          & 0.400          & \textbf{0.133} \\
Scheme & 0.151          & 0.377          & \textbf{0.642} & 0.360          & 0.191          \\ \hline
\end{tabular}%
}
\caption{Argument generation automatic evaluation results with best model highlighted for each metric.}
\label{tab:arg-gen-auto-eval}
\end{table}
We share our results in Table \ref{tab:arg-gen-auto-eval} and observe that compared to others, ArgU-Dual majorly yields better BLEU and RougeL scores and attains the best entailment results, indicating a better correlation with the original argument. On the contrary, using only argument schemes and stance-based control codes generally performs worse. We also observe that ArgU-Mono performs almost at par with ArgU-Stance across all metrics, whereas ArgU-Scheme contradicts the original argument the most. The results not only indicate the benefit of using both stance and scheme-based control codes but also indicate the superiority of the Dual architecture compared to Mono. 

\begin{table*}[t]
\resizebox{\textwidth}{!}{%
\begin{tabular}{l|l|l|l|l|l|l|l}
\hline
\multicolumn{1}{c|}{\textbf{ID}} &
  \multicolumn{1}{c|}{\textbf{Topic}} &
  \multicolumn{1}{c|}{\textbf{Variables}} &
  \multicolumn{1}{c|}{\textbf{Scheme}} &
  \multicolumn{1}{c|}{\textbf{Stance}} &
  \multicolumn{1}{c|}{\textbf{Argument Template}} &
  \multicolumn{1}{c|}{\textbf{Argument Text}} &
  \multicolumn{1}{c}{\textbf{Comments}} \\ \hline
1 &
  \multirow{4}{*}{\begin{tabular}[c]{@{}l@{}}Death \\      Penalty\end{tabular}} &
  \multirow{4}{*}{\begin{tabular}[c]{@{}l@{}}\textless{}VAR\_0\textgreater \\      human rights \\      around\\      the world \\      \textless{}VAR\_1\textgreater \\      mandatory \\      death sentence\end{tabular}} &
  \multirow{2}{*}{\begin{tabular}[c]{@{}l@{}}From \\      Source \\      Authority\end{tabular}} &
  Pro &
  \begin{tabular}[c]{@{}l@{}}\textless{}VAR\_0\textgreater supporters of the   bill \\      say it is a step toward \textless{}VAR\_1\textgreater{}\end{tabular} &
  \begin{tabular}[c]{@{}l@{}}human rights supporters of the   bills say it is\\      a step towards a mandatory death sentence\end{tabular} &
  \multirow{4}{*}{\begin{tabular}[c]{@{}l@{}}Generated \\      arg incorporates\\      input control \\      codes, variables \\      and generated \\      arg template\end{tabular}} \\ \cline{1-1} \cline{5-7}
2 &
   &
   &
   &
  Con &
  \begin{tabular}[c]{@{}l@{}}\textless{}VAR\_0\textgreater advocates have   long \\      argued that \textless{}VAR\_1\textgreater{}\end{tabular} &
  \begin{tabular}[c]{@{}l@{}}human rights advocates have long   advocated that \\      mandatory death sentences should be abolished\end{tabular} &
   \\ \cline{1-1} \cline{4-7}
3 &
   &
   &
  \multirow{2}{*}{\begin{tabular}[c]{@{}l@{}}Rule or\\      Principle\end{tabular}} &
  Pro &
  \begin{tabular}[c]{@{}l@{}}\textless{}VAR\_1\textgreater is not a violation   of \\      \textless{}VAR\_0\textgreater{}\end{tabular} &
  \begin{tabular}[c]{@{}l@{}}mandatory death sentence is not   a violation \\      of human rights\end{tabular} &
   \\ \cline{1-1} \cline{5-7}
4 &
   &
   &
   &
  Con &
  \begin{tabular}[c]{@{}l@{}}\textless{}VAR\_1\textgreater is a violation   \\      of \textless{}VAR\_0\textgreater{}\end{tabular} &
  \begin{tabular}[c]{@{}l@{}}mandatory death sentence is a   violation to \\      international human rights law\end{tabular} &
   \\ \hline
5 &
  \multirow{4}{*}{Abortion} &
  \multirow{4}{*}{\begin{tabular}[c]{@{}l@{}}\textless{}VAR\_0\textgreater \\      reproductive \\      health and rig-\\      hts advocates \\      \textless{}VAR\_1\textgreater \\      stop people \\      from having \\      abortions\end{tabular}} &
  \multirow{2}{*}{\begin{tabular}[c]{@{}l@{}}From \\      Conse-\\      quence\end{tabular}} &
  Pro &
  \begin{tabular}[c]{@{}l@{}}\textless{}VAR\_1\textgreater is an important   \\      step toward \textless{}VAR\_0\textgreater{}\end{tabular} &
  \begin{tabular}[c]{@{}l@{}}banning abortion is an important   stepping \\      toward reproductive rights\end{tabular} &
  \multirow{2}{*}{\begin{tabular}[c]{@{}l@{}}Pro \& con args \\      swapped\end{tabular}} \\ \cline{1-1} \cline{5-7}
6 &
   &
   &
   &
  Con &
  \begin{tabular}[c]{@{}l@{}}\textless{}VAR\_1\textgreater does nothing   \\      to \textless{}VAR\_0\textgreater{}\end{tabular} &
  \begin{tabular}[c]{@{}l@{}}banning abortion does nothing to   advance \\      women s reproductive rights\end{tabular} &
   \\ \cline{1-1} \cline{4-8} 
7 &
   &
   &
  \multirow{2}{*}{\begin{tabular}[c]{@{}l@{}}From \\      Source \\      Knowledge\end{tabular}} &
  Pro &
  \begin{tabular}[c]{@{}l@{}}\textless{}VAR\_1\textgreater has been proven   \\      to be effective in \textless{}VAR\_0\textgreater{}\end{tabular} &
  \begin{tabular}[c]{@{}l@{}}restricting access to abortion   has been proved \\      to be ineffective in protecting women s \\      reproductive rights\end{tabular} &
  \multirow{2}{*}{\begin{tabular}[c]{@{}l@{}}Generated \\      arg template \\      modified during \\      arg generation\end{tabular}} \\ \cline{1-1} \cline{5-7}
8 &
   &
   &
   &
  Con &
  \begin{tabular}[c]{@{}l@{}}\textless{}VAR\_1\textgreater is not the answer   to \\      \textless{}VAR\_0\textgreater{}\end{tabular} &
  \begin{tabular}[c]{@{}l@{}}banning abortion is not the   solution to \\      women s reproductive rights\end{tabular} &
   \\ \hline
\end{tabular}%
}
\caption{ArgU Generated Samples.}
\label{tab:arg-gen-examples}
\end{table*}

\subsubsection{Human Evaluation Results}
We perform a human evaluation study using the evaluators from Section \ref{human-annotators}. We created a worksheet with 50 random examples from the test set, where an example constitutes the argument topic, input KB variables, desired stance and argument scheme, the original argument from the dataset, and the generated argument text from each of the four models. The evaluators were asked to rate each generated argument text on the following five metrics. (i) \textbf{Fluency}: On a scale of 1 (low) to 5 (high), this scores the fluency and grammatical correctness of an argument. (ii) \textbf{Stance Appropriateness (Stance)}: On a binary scale, this determines if the stance exhibited by a generated argument aligns with the desired stance passed as control code. (iii) \textbf{Scheme Appropriateness (Scheme)}: On a binary scale, this determines if the argument scheme exhibited by a generated argument aligns with the desired scheme passed as control code. (iv) \textbf{Fact Faithfulness (Fact)}: On a scale of 1 (low) to 5 (high), this determines how well the generated argument incorporates the input variables. Ignoring variables or including additional facts (hallucination) are penalized. (v) \textbf{Logical Coherence (Logic)}: A subjective metric that rates the overall sensibleness of the logic portrayed by the generated argument text on a scale of 1 (low) to 5 (high). 

\begin{table}[h]
\resizebox{\columnwidth}{!}{%
\begin{tabular}{r|lllll}
\hline
\multicolumn{1}{c|}{\textbf{Model}} &
  \textbf{\begin{tabular}[c]{@{}c@{}}Fluency\\ (K=0.61)\end{tabular}} &
  \textbf{\begin{tabular}[c]{@{}c@{}}Stance\\ (K=0.87)\end{tabular}} &
  \textbf{\begin{tabular}[c]{@{}c@{}}Scheme\\ (K=0.9)\end{tabular}} &
  \textbf{\begin{tabular}[c]{@{}c@{}}Fact\\ (K=0.68)\end{tabular}} &
  \textbf{\begin{tabular}[c]{@{}c@{}}Logic\\ (K=0.71)\end{tabular}} \\ \hline
Mono   & \textbf{4.99} & 0.78*         & \textbf{0.83} & \textbf{3.89} & 4.01       \\
Dual   & 4.86        & 0.80*          & \textbf{0.83} & 3.88          & 4.06          \\
Stance & 4.95        & \textbf{0.84} & 0.79*          & 3.85          & 3.98*         \\
Scheme & 4.98          & 0.65*         & 0.79*          & 3.81          & \textbf{4.17} \\ \hline
\end{tabular}%
}
\caption{Argument generation human evaluation results with best model highlighted for each metric.}
\label{tab:arg-gen-human-eval}
\end{table}

We measure inter-annotator agreement by computing Cohens kappa (K) and observe substantial to high agreement across all metrics. Table \ref{tab:arg-gen-human-eval} shares the averaged ratings from both evaluators. For each metric, we highlight in bold the best performing model(s) and mark with an asterisk the model(s) where the difference from the best is at least 5\%. The fluency and fact metric results indicate that all models are fluent in generating arguments while incorporating the input variables, with ArgU-Mono performing the best. Trained with only stance-based control codes, ArgU-Stance yields the best results for stance appropriateness, while trained with only scheme-based control codes, ArgU-Scheme rates the lowest. Contrastly, ArgU-Scheme attains the highest rating for generating logically coherent arguments, whereas ArgU-Stance achieves the lowest rating. Thus, indicating the usefulness of using stance and scheme-based control codes for argument text generation. The ArgU-Dual and Mono variants rate similarly for both metrics, and rate high for scheme appropriateness, indicating that using control codes, the stance and scheme of an argument can be successfully controlled in tandem.

\subsection{Discussion}
Table \ref{tab:arg-gen-examples} contains arguments generated by ArgU-Dual. Examples 1 and 2 show the model's capability of generating authoritative argument text with the correct stance by referring to human rights advocates and supporters, thus exhibiting the ``From Source Authority'' argument scheme. Similarly, examples 3 and 4 denote the model's capability of generating appropriate argument text following the ``Rule or Principle'' argument scheme for both stances. Examples 5 and 6 depict a scenario where the generator demonstrates shallow understanding and inanely combines the input variables, yielding contrasting stance arguments. Examples 7 and 8 highlight cases where the argument decoder modifies the generated argument template, which in example 7 changes the meaning of the argument.

\section{Conclusion}
Here we propose ArgU: A neural factual argument generator that systematically generates arguments following a specified stance and argument scheme. We devise a multi-step annotation framework to yield two golden and silver standard annotated datasets that we further use to train multiple ArgU variants. Implementing automatic and human evaluation, we thoroughly analyze ArgU's generation capabilities. Our findings indicate ArgU's applicability for aiding users to formulate situation-specific arguments by controlling the argument stance and scheme using control codes.

\section*{Limitations}
As depicted in Table \ref{tab:arg-gen-examples}, there are scenarios where ArgU demonstrates a lack of understanding and instead paraphrases the input variables to generate an incorrect response. It seems likely that the model associates negation with Con. However, in examples 5 and 6, the model does not factor the word ``stop'' in Variable 1, leading to arguments that contradict the intended stance. Further, in examples 7 and 8, the argument decoder seems to modify the generated template, which changes the overall meaning of example 7. Such scenarios might reduce the trust in the model, hurting its practical use. 

All experiments involving ArgSpan, ArgSpanScheme, and ArgU only pertain to abortion, minimum wage, nuclear energy, gun control, the death penalty and school uniform. The model performance on any other topics is unknown. Although we test ArgSpanScheme on out-of-domain test sets, it still confines the six topics. Since ArgU is trained only on argument sentences with less than 150 tokens, it is more geared towards generating shorter arguments of less than 50 tokens. We further do not benchmark ArgU's inference time for practical use. 

\section*{Ethics Statement}
We acknowledge that all experiments were performed ethically and purely from an academic point of view. Although this research revolves around arguments from six sensitive topics, the argument generators were not explicitly trained to be discriminatory, exhibit bias, or hurt anyone's sentiments. Further, any generated text does not reflect the stance of the authors. The human evaluators were appointed and compensated as per the legal norms.

\bibliography{anthology,custom}
\bibliographystyle{acl_natbib}

\appendix

\section{Appendix}
\label{sec:appendix}

\begin{figure*}[h]
  \centering
  \includegraphics[width=\textwidth]{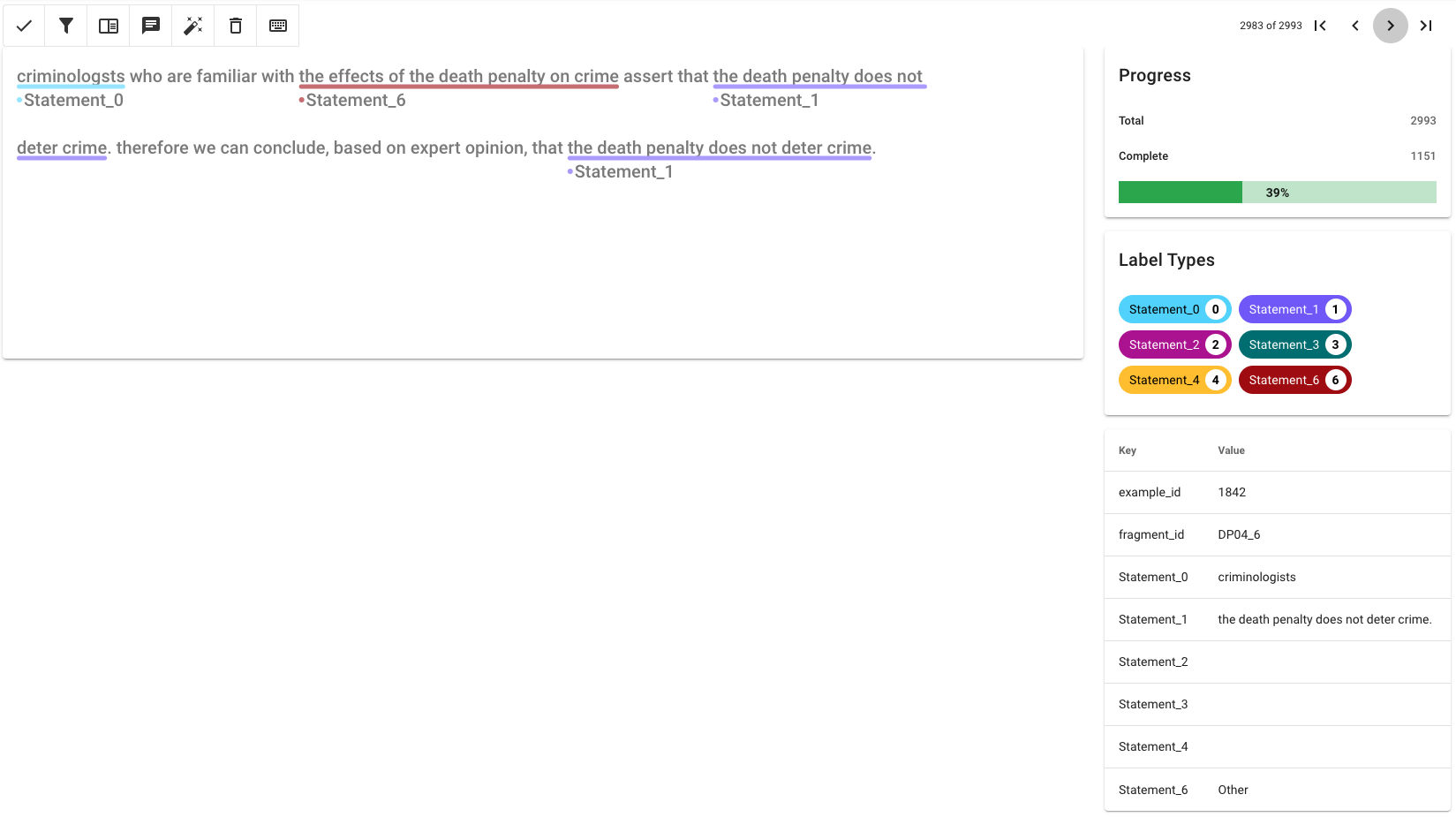}
  \caption{Doccano Annotation Screenshot.}
  \label{fig:doccano_screenshot}
\end{figure*}

\begin{figure}[h]
  \centering
  \includegraphics[width=\columnwidth]{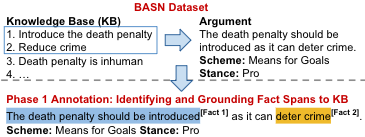}
  \caption{Phase 1 Annotation Pipeline.}
  \label{fig:phase_1}
\end{figure}

\begin{figure}[h]
  \centering
  \includegraphics[width=\columnwidth]{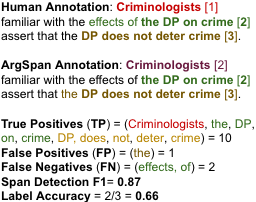}
  \caption{ArgSpan Evaluation.}
  \label{fig:ArgSpan-eval}
\end{figure}

\begin{figure}[h]
  \centering
  \includegraphics[width=\columnwidth]{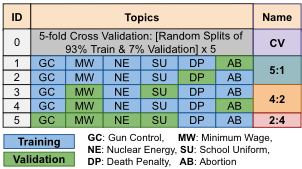}
  \caption{ArgSpanScheme Data Splits.}
  \label{fig:ArgSpanScheme-data}
\end{figure}

\begin{table*}[t]
\resizebox{\textwidth}{!}{%
\begin{tabular}{l|l|l|l}
\hline
ID &
  \multicolumn{1}{c|}{\textbf{Argument}} &
  \multicolumn{1}{c|}{\textbf{\begin{tabular}[c]{@{}c@{}}Actual \\ Label\end{tabular}}} &
  \multicolumn{1}{c}{\textbf{\begin{tabular}[c]{@{}c@{}}Annotator \\ Label\end{tabular}}} \\ \hline
1 &
  \begin{tabular}[c]{@{}l@{}}abortion is   necessary, because, unintended pregnancies are associated with birth defects,   \\ increased risk of child abuse, ad so on.\end{tabular} &
  \begin{tabular}[c]{@{}l@{}}means for \\ goal\end{tabular} &
  \begin{tabular}[c]{@{}l@{}}from \\ consequence\end{tabular} \\ \hline
2 &
  \begin{tabular}[c]{@{}l@{}}most students do not believe that school   uniforms are useful, so uniforms should not \\ be required.\end{tabular} &
  \begin{tabular}[c]{@{}l@{}}from source\\  knowledge\end{tabular} &
  \begin{tabular}[c]{@{}l@{}}from source \\ authority\end{tabular} \\ \hline
3 &
  \begin{tabular}[c]{@{}l@{}}the death penalty is unacceptable because   of the racial bias in the criminal justice system. \\ the death penalty does not   follow a fair criminal justice system because of its racial bias.\end{tabular} &
  \begin{tabular}[c]{@{}l@{}}rule or \\ principle\end{tabular} &
  \begin{tabular}[c]{@{}l@{}}from source \\ authority\end{tabular} \\ \hline
4 &
  \begin{tabular}[c]{@{}l@{}}it is not necessary to require school   uniforms, because t is important to respect students \\ who believe that school   uniforms are not necessary.\end{tabular} &
  \begin{tabular}[c]{@{}l@{}}from source \\ authority\end{tabular} &
  \begin{tabular}[c]{@{}l@{}}from source \\ knowledge\end{tabular} \\ \hline
5 &
  \begin{tabular}[c]{@{}l@{}}increasing the minimum wage reduces   income inequality. reducing income inequality \\ is desirable. we should   increase the minimum wage.\end{tabular} &
  \begin{tabular}[c]{@{}l@{}}from \\ consequence\end{tabular} &
  \begin{tabular}[c]{@{}l@{}}means for \\ goal\end{tabular} \\ \hline
\end{tabular}%
}
\caption{Annotator scheme conflicts}
\label{tab:annotator-scheme-conflict}
\end{table*}

\begin{figure*}[t]
  \centering
  \includegraphics[width=\linewidth]{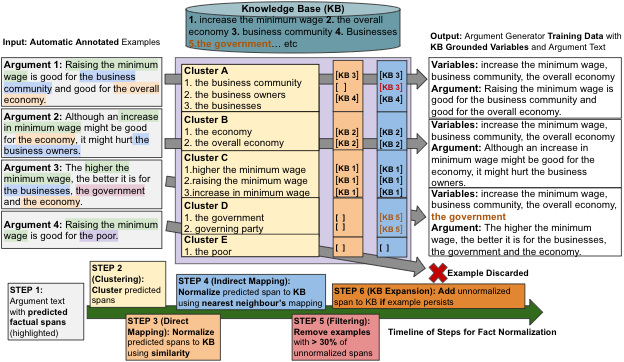}
  \caption{Phase 2 Dataset Fact Normalization Step.}
  \label{fig:fact_normalization}
\end{figure*}

\begin{table}[h]
\resizebox{\columnwidth}{!}{%
\begin{tabular}{r|l}
\hline
\multicolumn{1}{c|}{\textbf{Description}} &
  \multicolumn{1}{c}{\textbf{Tokens}} \\ \hline
\begin{tabular}[c]{@{}r@{}}Argument scheme \\ based control codes\end{tabular} &
  \begin{tabular}[c]{@{}l@{}}\textless{}from\_consequence\textgreater{},   \\ \textless{}from\_source\_authority\textgreater{}, \\ \textless{}from\_source\_knowledge\textgreater{},  \\ \textless{}goal\_from\_means/means\_for\_goal\textgreater{}, \\ \textless{}rule\_or\_principle\textgreater{}\end{tabular} \\ \hline
\begin{tabular}[c]{@{}r@{}}Argument  stance \\ based control codes\end{tabular} &
  \textless{}pro\textgreater{}, \textless{}con\textgreater{} \\ \hline
Variable   identifiers &
  \begin{tabular}[c]{@{}l@{}}\textless{}VAR\_0\textgreater{},   \textless{}VAR\_1\textgreater{}, \\ \textless{}VAR\_2\textgreater{}, \textless{}VAR\_3\textgreater{}\end{tabular} \\ \hline
Decoder BOS tokens &
  \textless{}pattern\textgreater{}, \textless{}argument\textgreater{} \\ \hline
\end{tabular}%
}
\caption{Special Tokens and Control Codes}
\label{tab:control-codes}
\end{table}

\subsection{ArgSpan Training Details}
\label{arg-span-details}
We initialize ArgSpan weights with pre-trained Roberta base weights, and train using 2 Nvidia RTX A5000 GPUs with mixed precision \cite{micikevicius2018mixed} and a batch size of 32. Prior to the biaffine layer, we reduce the hidden representation to 600 dimensions. We use a learning rate of 1E-5 and train till the validation loss stops improving for five steps. We also clip \cite{clipping} the gradients to a unit norm and use AdamW \cite{adamw} with the default PyTorch parameters for optimization.

\subsection{ArgSpanScheme Training Details}
\label{arg-span-scheme-details}
We initialize ArgSpanScheme weights with pre-trained Roberta base weights, and train using 1 Nvidia RTX A5000 GPUs with mixed precision and a batch size of 64. We use 2 layers of multi-headed self attention using 4 attention heads.  We use a learning rate of 1E-5 and train till the validation loss stops improving for five steps. We also clip the gradients to a unit norm and use AdamW with the default PyTorch parameters for optimization.

\subsection{ArgU Training Details}
\label{argu-details}
We initialize model weights with pre-trained BART \cite{lewis-etal-2020-bart} base weights and expand the embedding layer to accommodate 13 new tokens, detailed in Table \ref{tab:control-codes} Appendix \ref{sec:appendix}. We train all models over 2 Nvidia RTX A5000 GPUs with mixed precision and a batch size of 24. We use a learning rate of 1E-5 and train till the validation loss stops improving for five steps. We also clip the gradients to a unit norm and use AdamW with the default PyTorch parameters for optimization. We use beam search for decoding with a beam length of 5, a maximum length of 50 tokens, and a penalty for trigram repetitions in the generated argument.

\end{document}